# 대규모 언어 모델을 활용하는 부적절 발언 탐지를 위한

# 추론 관점을 명시하는 소프트 귀납적 편향 접근법

김주영[O1], 박지홍[1], 이세연[1], 박수진[2], 김건우[1]*
[1]경상국립대학교 컴퓨터공학과, [2]경상국립대학교 경영정보학과
{wndudwkd003, hong_0002, 2023014356, park_11412, gunwoo.kim}@gnu.ac.kr

# Soft Inductive Bias Approach via Explicit Reasoning Perspectives

# in Inappropriate Utterance Detection Using Large Language Models

Ju-Young Kim[O1], Ji-Hong Park[1], Se-Yeon Lee[1], Sujin Park[2], Gun-Woo Kim[1]*
[1]Dept. of Computer Science and Engineering, Gyeongsang National University
[2]Dept. of Management Information Systems, Gyeongsang National University

### 요 약

익명성이 보장된 일부 온라인 게임이나 커뮤니티에서는 무분별한 부적절 발언이 언어폭력과 범죄로 이어지는 사례가 속출하여 사회적 논쟁으로 대두되고 있으며, 건전한 소통 환경을 구축하기 위해 발화 문장에서 부적절 발언을 탐지하는 기법에 대응하는 연구의 필요성이 요구된다. 최근, 한국어 말뭉치를 학습하여 자연스러운 한국어를 구사하는 대규모 언어 모델과 생각의 연결고리 기법이 떠오르고 있지만, 부적절 발언 탐지 연구에서는 관련된 연구가 부족한 실정이다. 이 연구에서는 합리적인 사고를 수행하고 추론 과정에서 발생하는 문제점을 방지하기 위해 결론 도출을 위한 사고 관점을 미리 정의하는 소프트 귀납적 편향을 제안한다. 또한, 제안된 방법을 적용하여 한국어 대규모 언어 모델을 미세 조정하고, 학습 방법에 따른 모델의 성능을 정량적으로 비교 분석함과 동시에 모델의 출력을 정성적으로 평가하였다. 실험 결과, Kanana-1.5 모델에서 평균 정확도 87.0046을 달성하여 단순 지도학습 대비 약 3.89% 개선된 성능을 보였다. 이는 제안된 방법이 대규모 언어 모델의 지식을 단순 모방하는 것을 넘어, 추론 관점의 제약을 통해 정밀하고 일관된 판단을 가능하게 하는 것을 시사하며, 이러한 방법이 부적절 발언 탐지에 효과적임을 입증한다.

주제어: 부적절 발언 탐지(혐오 표현 탐지), 대규모 언어 모델, 생각의 연결고리, 추론 증류

## 1. 서론

인간은 언어를 사용하여 사람과 사람 간의 대화를 통해 복잡한 사고와 감정을 표현하고 소통하는 존재라는 것은 이미 널리 알려진 사실이다. 그러나, 일부 온라인 게임이나 커뮤니티의 소통 과정에서 익명성이라는 특수한 조건이 더해져 무분별한 부적절 발언이 발생하고 있다 [1,2]. 이러한 현상은 당사자 간의 갈등을 넘어 언어 폭력 그리고 심각한 범죄로 이어지는 사례[1]가 속출하여 사회적 논쟁거리로 떠오르고 있다. 따라서 대화 간의 부적절한 발언으로 인한 피해를 방지하기 위해 발화 문장에서 차별, 혐오, 비도덕 등의 표현을 탐지하여 건전한 소통 환경을 구축하는 부적절 발언 탐지 기법에 대한 심도 있는 연구가 요구된다[3,4].

최근에는 자가-주의 집중 기술을 활용한 트랜스포머[5]의 성공을 시작으로 다양한 대규모 언어 모델이 등장하였으며, 한국어 말뭉치로 학습하여 자연스러운 한국어를 구사하는 대규모 언어 모델이 연구되고 있다[6,7,8]. 이러한 대규모 언어 모델과 관련하여, 혁신적인 지시 기법으로 제안된 생각의 연결고리[9]는 기존의 입력 질문에 대해 바로 최종 답변을 생성하는 방법 대신, 중간 추론 과정을 통해서 복잡한 문제를 단계적으로 사고하도록 지시하여 최종 답변을 유도한다. 그러나, 한국어 기반의 대규모 언어 모델의 발전과 생각의 연결고리 기법의 혁신에도 불구하고, 이를 활용한 한국어 중심의 부적절 발언 탐지 연구는 미비한 실정이다.

이 연구에서는 한국어 말뭉치로 학습된 대규모 언어 모델과 생각의 연결고리를 활용하여 발화자 간의 대화에서 부적절한 발언을 탐지하는 모델을 개발한다. 특히, 합리적인 사고를 수행하기 위해 중간 추론 과정을 GPT-4.1[10]로부터 답변을 추출하고, 결론에 도달하기 위한 사고 관점을 미리 정의하는 소프트 귀납적 편향을 제안한다. 논문에서 제안하는 방법을 통해 대규모 언어 모델을 미세 조정하여 사전 학습된 모델과 추론 관점에 따른 부적절 발언 탐지 성능을 정량적으로 평가하고, 데이터에 따른 모델의 생성 결과를 정성적으로 분석하여 대규모 언어 모델과 생각의 연결고리를 활용하는 부적절 발언 탐지 연구에 기여하고자 한다.

## 2. 관련 연구

---

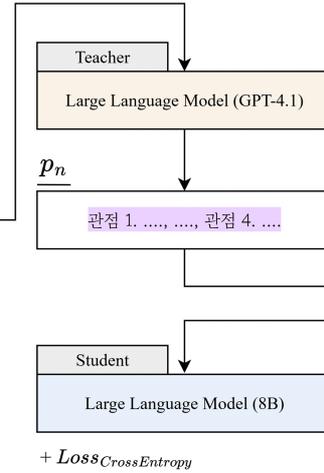

그림 1. 이 그림은 부적절 발언 탐지를 위해 논문에서 제안하는 추론 관점을 명시하는 소프트 귀납적 편향을 적용하여 GPT-4.1 대규모 언어 모델로부터 추론 증류를 수행하고 한국어 기반의 대규모 언어 모델을 미세조정 과정을 나타낸다.

기존의 한국어를 중심으로 하는 부적절 발언 탐지 연구에서는 기계학습과 자연어처리(NLP) 기술을 결합하여 부적절 발언을 탐지하고자 하였다[11,12]. 그러나 한국어는 어근에 다양한 접사, 어미, 조사가 결합하여 수많은 불규칙 형태로 활용되며, 이러한 교착어 특성으로 인해 한국어 처리의 근본적인 한계가 존재한다[13].

원초적으로 인간의 언어 표현은 미묘하고 복잡한 함축적 의미를 포함하고 있으므로, 딥러닝과 대규모 언어 모델을 활용하여 상황 맥락을 이해하는 고도화된 모델로 개발하는 것이 필수적이다. 선행 연구에서는 KoBERT[14], KoELECTRA[15] 기반의 모델을 활용한 연구가 이뤄졌으나 모델의 구조적 한계로 인해 대화의 길이가 짧고, 전체 맥락을 충분히 반영하지 못하는 문제점이 있다.

생각의 연결고리[9] 기법은 명시적인 학습 없이 지시문을 수정하는 방법으로 제안되었으나, 부적절 발언 탐지와 같은 특화된 다운스트림 작업에서 효과적으로 활용하기 위해서는 해당 도메인에 적합한 추론 패턴이 포함된 데이터로 미세 조정하는 과정이 필수적이다[16].

지식 증류 기법은 교사 모델과 학생 모델이 존재하며, 일반적으로 교사 모델이 학생 모델보다 더 큰 파라미터로 학습된다. 그리고 교사 모델의 지식을 학생 모델로 전달하여 학습하는 패러다임을 갖고 있다. 이러한 배경속에서 GPT와 같은 대규모 언어 모델로부터 Few-shot CoT (Chain of Though) 또는 Zero-shot CoT, 즉 질문과 답변을 입력하여 사고 과정을 생성하여 소형 모델에 학습하는 추론 증류[17,18] 기법이 등장하였다.

하지만, 추론 과정에 대하여 명시적인 관점을 제시하지 않고 단순히 모델이 생성한 추론 과정을 그대로 학습하는 경우, 일관된 사고의 방향성 제약, 즉 귀납적 편향이 미약하여 방대한 추론 경로로 이어지는 문제점과 더

불어 모델 간 능력 격차로 인한 표면적 모방, 중간 추론 과정의 오류 전파, 그리고 도메인 특화 맥락의 부재 등의 한계가 발생할 여지가 있다.

## 3. 연구 방법론

### 3.1. 대규모 언어 모델을 활용한 추론 증류

이 연구에서는 부적절 발언을 탐지하기 위해 대략 80억(이하, 8B) 정도의 파라미터를 갖고 있는 한국어 기반의 대규모 언어 모델을 개발한다. 그러나, 단순히 지도학습을 수행하였을 때 종종 학습이 의도치 않게 이뤄지는 경험적 결론이 도출되었다. 이는 8B 정도의 모델은 질문과 답변으로 이뤄진 일반적인 지도학습만으로는 한계가 있으며, 전체 대화가 주어지더라도 전반적인 맥락을 직관적으로 이해하지 못할 뿐만 아니라 문장에서 드러나는 인간의 미묘한 감정을 포착하여 부적절 발언을 탐지하는 과정에 어려움이 있다는 것이 시사된다.

이러한 문제점을 해결하기 위해, 이 연구에서는 대규모 언어 모델을 교사 모델로 하여 질문-정답 쌍을 입력하고 중간 사고 과정을 추출하며, 이를 학생 모델의 미세조정에 활용하는 추론 증류 방법으로 모델 개발을 수행한다. 추출된 중간 사고 과정은 텍스트 앞머리에 '<think></think>' 태그로 감싸지며, 기존의 정답은 '<answer></answer>' 태그로 감싸 완성된다. 이 태그 표현 방법은 DeepSeek-R1[16] 에서 기반하였으며, 추론 과정을 캡슐화하여 중간 사고 과정과 답변을 명확히 구분하고, 모델의 설명 가능성을 확장하는 장점이 있다.

그러나, GPT-4.1과 같은 정확한 파라미터와 구조가 공개되지는 않았지만, 대략 조 이상 단위의 파라미터로 대규모 언어 모델 중에서도 초거대 구조를 가질 것으로 추정되는 모델의 사고 과정을 그대로 학습하게 되면 모델의 구조적 차이로 인해 복잡한 추론 과정을 재현하지 못

하는 문제가 발생할 수 있다. 또한, GPT-4.1의 답변을 추출하는 과정에서 특수한 조건이 없다면 수많은 추론 경로로 이어지거나, 또는 단순히 중복된 사고 과정이 추출될 수 있다.

### 3.2. 관점 제약 중심의 소프트 귀납적 편향

귀납적 편향은 인공신경망의 층 개수 또는 합성곱 신경망의 지역적 연결성과 같은 모델의 구조적 제약 등을 의미하며, 모델이 학습되지 않은 데이터에 대해 일반화할 때 특정한 가정이나 선호를 갖도록 한다[19]. 이 연구에서는 추론 증류를 통해 대규모 언어 모델을 개발함과 동시에 사고 과정에 특정 관점을 명시하는 소프트 귀납적 편향을 제안한다.

하드 귀납적 편향은 고정적 특성과 같은 구조적 제약에 집중하는 반면, 이 논문에서 제안하는 소프트 귀납적 편향은 교사 모델과 학생 모델이 정답에 도출하기 위한 명시적인 관점을 제시하여 사고 과정에 대한 제약을 통해 정확한 답변을 유도한다. 제시된 관점은 표 1과 같이 표면적 · 원인적 · 영향적 · 종합적으로 총 4가지이다. 각 관점에 대한 지시문은 교사 모델의 답변을 추출하는 과정과 학생 모델을 학습할 때 본래의 지시문과 함께 입력된다. 여기서 본래의 지시문은 전체 대화와 분석 대상 발화문을 포함하는 지시문을 의미한다.

표 1. 사고 관점 명시를 위한 지시문

| 사고 관점 | 구체적인 관점 지시문 |
| --- | --- |
| 표면적 | 대화에서 사용된 언어 표현 중 부적절한 단어나 표현(욕설, 비속어)이 발견되었는지? |
| 원인적 | 대화가 부적절한 발화의 경우 어떤 원인(차별, 혐오, 편향, 폄하 등)에서 문제가 되는지? |
| 영향적 | 대화의 흐름을 고려했을 때 각 발화자의 발언이 대화 당사자에게는 어떤 영향을 주고, 제3자에게는 어떤 영향을 미치는지? |
| 종합적 | 대화의 전체 맥락과 발화 간의 상호작용을 고려하여 왜 각 발화가 적절 또는 부적절로 판단되었는지? |

먼저, *표면적 관점*은 발화 문장의 표면에서 부적절 단어가 존재하는지 관찰하는 것이다. 트랜스포머 기반의 모델은 서브워드 토크나이제이션을 통해 Out-of-Vocabulary(OOV) 문제를 해결할 수 있어 표면에서 드러나는 부적절 발언을 탐지하는데 용이하다. 이 관점은 해당 단어나 표현이 왜 사용되었는지에 대한 판단 없이 순수하게 표면적인 의미만을 분석한다. *원인적 관점*은 부적절한 발화가 감지된 경우, 왜 그러한 표현이 사용되었는지 근본적인 원인을 탐구한다. 차별, 혐오 등의 동기나 의도를 분석하여 부적절 발언의 배경을 이해하고자 한다. *영향적 관점*은 발언이나 단어로 인해 대화 상대방 또는 제3자에게 미치는 영향을 판단한다. 발언이 상대방에게 어떤 감정적, 심리적 영향을 주는지, 사회적 맥락에서 어떤 파급효과를 가져올 수 있는지 고려한다. 마지막으로

*종합적 관점*은 앞선 3가지 관점이 모두 수동으로 설정되었기 때문에 놓칠 수 있는 부분을 대규모 언어 모델이 종합적으로 고려한다. 이는 이전에 정의된 관점이 적절하게 고려되면 자연스럽게 일관된 사고를 통한 올바른 추론 경로로 이어질 것으로 가정한다.

결론적으로, 교사 모델인 GPT-4.1 모델과 적은 파라미터의 수를 가진 작은 학생 모델의 추론 경로를 명시하는 제약은 질문으로부터 정확한 정답으로 도출하기 위한 이정표와 같은 역할을 하게 되며, 큰 모델의 지식을 증류하여 적은 파라미터의 모델이더라도 큰 모델의 성능을 기대할 수 있다. 논문에서 제안하는 방법은 그림 1과 같이 단계 1, 단계 2로 나누어 수행하며 부적절 발언을 탐지하기 위한 한국어 대규모 언어 모델 학습을 진행한다.

## 4. 실험 및 평가

### 4.1. 데이터 소개

이 연구에서는 부적절 발언 탐지 모델을 개발하기 위해 국립국어원 AI말평 대화 맥락 기반 부적절 발언 탐지 과제에서 제공하는 데이터 세트를 사용한다. 이 데이터 세트는 학습, 검증, 평가로 나뉘어 제공되며, 학습과 검증 세트는 정답이 포함되어 있지만, 평가 세트는 정답이 제공되지 않아 누리집에서 제공하는 평가 기능으로 산출된 점수에 의존한다. 데이터는 그림 1과 각 발화에 '적절', '부적절'의 정답 쌍으로 구성되며, 2인의 발화자가 서로 번갈아 가며 대화가 이뤄진다. 또한, 각 발화자에 대한 메타 정보가 존재하지 않으므로 익명성을 가진다. 따라서 모델은 대화에 의존하여 부적절 발언을 탐지하게 된다. 실제 모델에 입력되는 데이터를 구성하고, 대화의 맥락을 고려하기 위해 데이터 세트를 전체 대화와 각각의 분석 대상 발화로 나누어 변환한다. 이때 변환 전후의 데이터 수는 표 2와 같다.

표 2. 데이터 세트 변환에 따른 데이터 수 (단위: 개)

| 구분 | 학습 | 검증 | 평가 |
| --- | --- | --- | --- |
| 변환 전 | 504 | 101 | 397 |
| 변환 후 | 2,109 | 410 | 1,657 |

### 4.2. 모델 학습

이 연구에서는 Hugging Face에 공개된 2개의 한국어 사전학습 모델을 기반으로 실험을 수행하였다[7,8]. 학습 방법은 Low-Rank Adaption(LoRA)[20] 기법을 활용하여 대규모 언어 모델을 효율적으로 미세조정 하였다. LoRA는 사전 학습된 가중치를 고정하고 저차원 분해 행렬을 추가하여 적은 수의 파라미터만으로도 효과적인 학습이 가능한 기법이다. 구체적인 수식으로 나타내면 LoRA의 가중치 업데이트 방식을 수식 1로, 순전파 과정을 수식 2로 나타낼 수 있다.

여기서 $W \in \mathbb{R}^{d \times k}$는 기존 가중치 행렬이고, $B \in \mathbb{R}^{d \times r}$과 $B \in \mathbb{R}^{r \times k}$는 학습 가능한 저차원 행렬이다. 또한, $d$는 각각 출력 벡터의 차원을, $k$는 입력 벡터의 차원을

의미한다. 이 연구에서는 표 3의 $S$에 속한 모듈에 대하여 LoRA를 적용하였고, 랭크와 스케일 하이퍼파라미터는 $r$과 $a$에 해당한다.

$$W' = W + \Delta W, \quad \Delta W = sBA, \quad s = \frac{a}{r} \quad (1)$$

$$h = Wx + sBAx \quad (2)$$

그리고 학습 과정에서 Unsloth[21] 프레임워크를 활용하여 메모리 효율성과 학습 속도를 향상했다. Unsloth은 최적화된 커널을 활용하여 LoRA 학습 시 역전파 과정을 가속화하고, 대략 2배 빠른 학습을 가능하게 한다. 최종적으로 대규모 언어 모델을 학습하는데 널리 알려진 교차 엔트로피 손실 함수를 사용하여 학습을 진행한다. 여기서 교차 엔트로피 손실 함수는 수식 3과 같다.

$$L = -\frac{1}{N} \sum_{i=1}^{N} \sum_{t=1}^{T^{(i)}} \log P(y_t^{(i)} \mid x^{(i)}, y_{<t}^{(i)}; \theta) \quad (3)$$

표 3. 모델 개발 환경

| 분류 | 환경 및 설정 |
|---|---|
| CPU | i9-10900X |
| GPU | GeForce RTX 3090 (24GB) |
| 운영체제 | Ubuntu 22.04 |
| CUDA | 12.6 |
| 파이썬 | 3.12.11 |
| PyTorch | 2.7.1 |
| 시드 | 42, 2025, 7412013 |
| Rank ($r$) | 64 |
| LoRA Alpha ($a$) | 64 |
| Dropout | 0.0 |
| Epoch | 5 |
| Metric for Best Model | Eval Loss |
| Learning Rate | 2e-5 |
| Train Batch Size | 1 |
| LR Scheduler | Cosine with Restarts |
| Target modules ($S$) | q_proj,k_proj,v_proj,o_proj, gate_proj,up_proj,down_proj |
| EXAONE-3.5 (8B) | LGAI-EXAONE/EXAONE-3.5 -7.8B-Instruct |
| Kanana-1.5 (8B) | kakaocorp/kanana-1.5-8b-base |

또한, 이 논문에서는 제안된 방법을 검증하기 위해 비교 모델을 구성하며, 일반적으로 학습 방법인 지도학습(SFT)과 명시적인 추론 관점 없이 추출된 CoT 지식을 증류하여 지도학습을 수행하는 방법(SFT with CoT)을 통해 비교한다. 두 방법 모두 LoRA와 Unsloth를 활용하며, 표 3의 하이퍼파라미터를 따른다.

### 4.3. 결과 분석

실험 결과에 대한 분석은 세 가지로 이뤄지며, 모델 학습 방법에 따른 정량적 평가, 추론 관점에 따른 정량적 평가, 모델의 출력 결과에 따른 정성적 평가 순서대로 진행한다. 먼저, 모델 학습 방법에 따른 정량적 평가이다. 학습 결과, 표 4와 같이 기존의 SFT 방법이 대부

분 단순 추출 CoT를 활용한 SFT 방법보다 더 높은 점수를 달성하였다. 이러한 결과는 단순히 CoT를 추출하고 학습하는 과정에서 불필요한 추론 경로가 발생하였거나, 학습에 사용된 8B 모델이 더 높은 파라미터를 가진 모델의 지식을 제대로 학습하지 못하는 것을 보여준다.

그러나, 제안하는 방법으로 학습하였을 때 EXAONE-3.5 모델에서는 평균 정확도 85.56을 Kanana-1.5 모델에서는 평균 정확도 87.0046을 달성하였으며, 기존 SFT 방법의 평균 정확도 84.6308, 83.7457보다 각각 1.0979%, 3.8917% 더 높은 성능을 달성하였다. 또한, SFT with CoT 방법에서 각각 평균 정확도 83.6854, 84.0274로 나타났으며, 제안하는 방법이 각각 2.239%, 3.544% 더 높은 결과가 나타났다. 이러한 정량적 평가로 미루어 보았을 때 논문에서 제안한 추론 관점을 명시하는 소프트 귀납적 편향이 모델의 추론 증류 과정에서 의미 있는 개선 효과를 불러일으킨다는 것을 시사한다.

표 4. 모델 학습 방법에 따른 부적절 발언 탐지 결과

| 모델 | 시드 | 부적절 발언 탐지 정확도 | | |
|---|---|---|---|---|
| | | SFT | SFT with CoT | Proposed |
| EXAONE-3.5 | 42 | **85.9988** | 83.1623 | 85.3953 |
| | 2025 | 83.5244 | 84.3090 | **86.1798** |
| | 7412013 | 84.3693 | 83.5848 | **85.0935** |
| | 평균 | 84.6308 | 83.6854 | **85.5562** |
| Kanana-1.5 | 42 | 82.4985 | 83.8262 | **87.1454** |
| | 2025 | 84.0072 | 83.4641 | **86.7230** |
| | 7412013 | 84.7314 | 84.7918 | **87.1454** |
| | 평균 | 83.7457 | 84.0274 | **87.0046** |

다음은 추론 관점에 따른 정량적 평가이다. 학습 방법에 따른 실험에서 세 가지 시드의 평균 정확도가 가장 높았던 Kanana-1.5 모델을 사용하여 제안하는 방법으로 학습하되, 관점 단계를 다르게 하여 학습하였다. 학습 결과, 표 5와 같이 4가지 관점을 전부 사용하였을 때 평균 정확도 87.0046을 달성하였으며, 이는 하나만 사용하는 경우보다 0.7451% 더 높은 결과이다. 이러한 결과는 부적절 발언 탐지를 위해 설계된 4가지 관점이 모델의 정확한 추론에 효과가 있다는 것을 나타낸다. 또한, 관점 단계 추가에 따른 상승 폭을 더 높이기 위해서는 부적절 발언과 관련된 인문학적 배경지식을 바탕으로 정교한 관점을 설계하는 과정이 필요할 것으로 사료된다.

표 5. 추론 관점에 따른 부적절 발언 탐지 결과

| 모델 | 시드 | 부적절 발언 탐지 정확도 | | | |
|---|---|---|---|---|---|
| | | 1단계 | 2단계 | 3단계 | 4단계 |
| Kanana-1.5 (Proposed) | S1 | 85.9988 | 86.7833 | 86.9644 | **87.1454** |
| | S2 | 86.1798 | 86.4816 | **87.3265** | 86.7230 |
| | S3 | 86.9040 | 85.5160 | 86.3609 | **87.1454** |
| | 평균 | 86.3609 | 86.2603 | 86.8839 | **87.0046** |

* S1: 42, S2: 2025, S3: 7412013

표 6. 검증 데이터 세트 보기

| 데이터 구분 | 발화 식별자 | 발화 내용 | 실제 정답 |
|---|---|---|---|
| 검증 데이터 | dev-00069-001 | 이대 중데보다는 쉽겠지 생각했다니 어머니가 대학 안 나오셨나 봐요 | 부적절 |
| | dev-00069-002 | 이런 글에 왜 이상한테 꼬인 댓글을 쓸까. 학벌에 열등감 있나.. | 부적절 |
| | dev-00069-003 | 님이 모자란 건데 왜 상대방을 탓해요? 인서울 다 힘들어요. | 부적절 |
| | dev-00069-004 | 아무튼 인성 자체가 삐뚤어진 사람들이 유난스럽게 열심히 댓글을 닮. | 부적절 |

표 7. 모델 학습 방법에 따른 부적절 발언 탐지 출력 결과

| 모델 구분 | 발화 식별자 | 모델 출력 | 최종 결과 |
|---|---|---|---|
| Kanana-1.5 (SFT) | dev-00069-003 | 적절 | 오답 |
| Kanana-1.5 (SFT with CoT) | dev-00069-003 | \<think\>1. 상대방을 비난하는 대신 자신의 생각을 전달하며, 논쟁의 감정적 대립을 완화(생략). 2. 상대방을 비난하거나 조롱하지 않고(생략). 3. (생략) 자신의 생각을 전달하며 대화의 흐름을 유지하고(생략). 4. (생략) 따라서 발화는 대화 맥락에서 적절합니다.\</think\>\<answer\>적절\</answer\> | 오답 |
| Kanana-1.5 (Proposed) | dev-00069-003 | \<think\>1. "님이 모자란 건데"라는 표현은 상대를 비하하는 비속어적 언어(생략). 2. 이 발화는 상대방을 "모자란" 사람으로 일반화(생략). 3. (생략) 대화 당사자 간 감정적 대립 (생략) 불쾌한 분위기를 조성할 수 있습니다. 4. P2의 인신공격적 발언에 대해 P1이 상대방을 비하하는 표현 (생략) 상호 존중이 필요한 대화 맥락에서 부적절합니다.\</think\>\<answer\>부적절\</answer\> | 정답 |

마지막으로 모델 출력 결과에 따른 정성적 평가이다. 기존의 평가 데이터는 정답이 주어지지 않기 때문에 평가 데이터가 아니라 검증 데이터 세트를 사용하여 실험을 진행한다. 표 6은 검증 데이터 세트에서 제안된 모델의 경우에만 정답을 맞힌 샘플을 보기로 설정하였다. 표 7은 학습 방법에 따른 모델의 출력 결과를 나타낸다. 실험 결과, 기존의 SFT는 단순히 최종 출력으로 곧바로 생성하므로 어떠한 중간 과정이 있는지 불분명한 특징이 드러난다. 또한, 단순 추출 CoT의 경우 초기 단계부터 오답을 마치 정답처럼 추론하여, 최종 결론 또한 오답으로 도출되었다. 그러나, 제안된 모델의 경우에는 중간 과정을 명확히 확인할 수 있는 것과 동시에 관점 단계마다 다른 모델보다 더 논리적이고 구체적인 예시를 들며, 각 관점 단계를 충실히 수행하여 최종 출력이 정답으로 도출되는 것을 확인할 수 있었다.

## 5. 결론

이 논문에서는 한국어 대규모 언어 모델을 기반으로, 부적절 발언 탐지 과제에서 추론 관점을 명시하는 소프트 귀납적 편향 접근법을 제안하였다. 기존의 교사 모델의 지식을 명시적인 관점 없이 추출하는 단순 추론 증류 기법의 한계 즉, 불필요한 추론 경로 발생, 성능 격차로 인한 단순 모방 등의 문제를 극복하기 위해 표면적·원인적·영향적·종합적 등 4가지 관점을 명시하여 추론 경로를 제약하였다. 또한, 제안하는 방법론을 검증하기 위해 EXAONE-3.5와 Kanana-1.5 모델을 기반으

로 모델의 학습 방법, 사고 관점 단계에 따른 두 가지 정량적 평가를 수행하고, 모델의 출력을 바탕으로 정성적 평가를 진행하였다. 실험 결과, 두 대규모 언어 모델에서 모두 기존 방법 대비 유의미한 성능 향상을 보였으며, 특히 Kanana-1.5 모델에서 평균 정확도 87.0046을 달성하여 단순 SFT 대비 약 3.8917% 개선 효과를 입증하였다. 또한, 추론 관점 단계를 모두 활용했을 때 가장 높은 평균 정확도를 달성하였으며, 정성적 평가에서도 제안된 방법이 중간 사고 과정을 명확히 제시하면서 논리적이고 구체적인 판단을 도출하였다. 이러한 결과는 대규모 언어 모델의 지식을 단순 모방하는 방법을 넘어, 합리적이고 일관된 추론 과정을 유도하는 귀납적 편향의 필요성을 확인할 수 있었다. 마지막으로, 이 연구에서는 한국어 대화 맥락에서 발생하는 부적절 발언 탐지 문제에서 추론 증류, 소프트 귀납적 편향 접근법을 제시하여 효과적인 부적절 발언 탐지 모델 구축에 기여한다. 향후 연구에서는 사고 관점을 스스로 구축하는 아키텍처를 설계하여 다운스트림 작업의 도메인 지식이 부족하더라도 높은 접근성을 위한 확장된 연구로 진행하고자 한다. 이를 통해 다양한 맥락과 응용 분야에서 활용 가능한 범용적 프레임워크로 발전할 것으로 기대한다.

## 사사문구

의 연구결과임.

# 참고문헌